%% file: 11904.tex
\documentclass[runningheads]{llncs}

\usepackage{eccv}

\usepackage{eccvabbrv}

\usepackage{graphicx}
\usepackage{booktabs}
\usepackage[misc]{ifsym}

\usepackage[accsupp]{axessibility}  

\usepackage{amsmath,amssymb}
\usepackage{mathrsfs}

\usepackage{multirow}
\usepackage{threeparttable}

\usepackage{hyperref}

\usepackage{orcidlink}

\newlength\savewidth\newcommand\shline{\noalign{\global\savewidth\arrayrulewidth
  \global\arrayrulewidth 1pt}\hline\noalign{\global\arrayrulewidth\savewidth}}
\newcommand{\tablestyle}[2]{\setlength{\tabcolsep}{#1}\renewcommand{\arraystretch}{#2}\centering\footnotesize}

\newcommand{\red}[1]{{#1}}

\newcommand\blfootnote[1]{%
  \begingroup
  \renewcommand\thefootnote{}\footnote{#1}%
  \addtocounter{footnote}{-1}%
  \endgroup
}

\newcommand{\nbf}[1]{{\noindent \textbf{#1}}}

\begin{document}

\title{FAFA: Frequency-Aware Flow-Aided Self-Supervision for Underwater Object Pose Estimation}

\titlerunning{FAFA for Self-Supervised Underwater Object Pose Estimation}

\author{Jingyi Tang\inst{1,2}$^*$\orcidlink{0000-0002-6681-2103} \and
Gu Wang\inst{1}$^*$\orcidlink{0000-0002-0759-0782} \and
Zeyu Chen\inst{1}\orcidlink{0009-0004-5418-8827} \and 
Shengquan Li\inst{2}\orcidlink{0000-0002-4504-3659} \and 
Xiu Li\inst{1,2}\textsuperscript{(\Letter)}\orcidlink{0000-0003-0403-1923} \and 
Xiangyang Ji\inst{1}\orcidlink{0000-0002-7333-9975}
}
\authorrunning{J.~Tang, G.~Wang et al.}

\institute{Tsinghua University \\
\email{\{tangjy21@mails.,wanggu1@,chen-zy22@mails.,li.xiu@sz.,xyji@\}tsinghua.edu.cn}\blfootnote{$*$ Equal contribution. \Letter~Corresponding author.} \and
Peng Cheng Laboratory \\
\email{lishq@pcl.ac.cn}
}

\maketitle

\input{sections/0_abstract}

\input{sections/1_intro}
\input{sections/2_related}
\input{sections/3_method}
\input{sections/4_exp}
\input{sections/10_conclusion}


%
%
\bibliographystyle{splncs04}
\bibliography{egbib}
\end{document}

%% file: sections/0_abstract.tex
\begin{abstract}

Although methods for estimating the pose of objects in indoor scenes have achieved great success, the pose estimation of underwater objects remains challenging due to difficulties brought by the complex underwater environment, such as degraded illumination, blurring, and the substantial cost of obtaining real annotations. In response, we introduce FAFA, a \textbf{F}requency-\textbf{A}ware \textbf{F}low-\textbf{A}ided self-supervised framework for 6D pose estimation of unmanned underwater vehicles (UUVs). Essentially, we first train a frequency-aware flow-based pose estimator on synthetic data, where an FFT-based augmentation approach is proposed to facilitate the network in capturing domain-invariant features and target domain styles from a frequency perspective.
Further, we perform self-supervised training by enforcing flow-aided multi-level consistencies to adapt it to the real-world underwater environment. Our framework relies solely on the 3D model and RGB images, alleviating the need for any real pose annotations or other-modality data like depths. We evaluate the effectiveness of FAFA on common underwater object pose benchmarks and showcase significant performance improvements compared to state-of-the-art methods. Code is available at \href{https://github.com/tjy0703/FAFA}{github.com/tjy0703/FAFA}.
\keywords{Underwater Object Pose \and Self-Supervision \and FFT \and Flow}
\end{abstract}

%% file: sections/1_intro.tex
\section{Introduction}
\label{sec:intro}

Estimating the 6D object pose for unmanned underwater vehicles (UUVs) has a wide range of applications, such as underwater vehicle tracking \cite{joshi2020deepurl}, underwater archaeology \cite{xanthidis2022multi}, marine resource development \cite{kalwa2012european}, and underwater intervention tasks \cite{sapienza2023model, casalino2016underwater}.
Currently, remarkable advancements have been made in the general domain of 6D object pose estimation, particularly in indoor scenarios \cite{li2019cdpn, wang2021gdr, hu2022perspective, Xu2022pose, sundermeyer2023bop,hodan2024bop}.
However, due to the complexity of the underwater environment, in most cases, these methods struggle to achieve satisfactory results when directly applied to underwater object pose estimation \cite{fan2022deep}.
Therefore, accurately estimating the 6D pose of objects in underwater environments remains a formidable challenge.

\input{figs/0_teaser}

The challenges in underwater object pose estimation are mainly threefold. 
Firstly, due to the complex optical effects in water, real underwater images typically exhibit greater degradation in terms of illumination conditions and clarity, resulting in a larger domain gap between them and synthetic images than in general scenes \cite{lu2017underwater}. 
Secondly, because near-infrared light is attenuated faster than visible light underwater, ordinary depth cameras oftentimes fail to work \cite{gonzalez2023survey}. 
Moreover, as it is more difficult to place additional calibration devices and the lack of depth data, obtaining real pose annotations for underwater objects is much more costly \cite{joshi2020deepurl,tang2023rov6d}.

To bypass the challenge of annotating real underwater data, previous works \cite{joshi2020deepurl,sapienza2023model} mainly focus on training with synthetic underwater data generated by Unreal Engine 4 (UE4) \cite{manderson2018aqua} or Unity3D \cite{borkman2021unity}.
However, networks trained on synthetic data often exhibit poor generalization on real data due to the sim2real gap. 
To improve the pose estimation accuracy, a commonly utilized approach is to refine the initial prediction with the render and compare paradigm \cite{li2020deepim,labbe2020cosypose,lipson2022coupled}. 
Though effective to some extent, such methods still suffer from notable domain discrepancies.
%
%
%
Recently, self-supervised object pose estimation methods \cite{chen2022sim,wang2020self6d, deng2020self,lin2022category,wang2021occlusion,hai2023pseudo} have demonstrated their effectiveness in mitigating the domain gap by leveraging unlabeled real images to enhance the synthetically pre-trained network. 
However, the absence of depth data in the underwater environment renders most of them 
 inapplicable \cite{deng2020self,wang2020self6d,chen2022sim,lin2022category,wang2021occlusion}.
\cite{hai2023pseudo} proposes to integrate strengths of both pose refinement and self-supervised learning via pseudo flow consistency, eliminating the need for depth.
Nevertheless, these methods typically perform self-training by treating the network as a black box, \ie, solely enforcing alignments with the outputs from the teacher and student networks while neglecting the impact of the feature space \cite{wang2020self6d,wang2021occlusion,hai2023pseudo}.


In this work, we propose FAFA, a two-stage \textbf{F}requency-\textbf{A}ware \textbf{F}low-\textbf{A}ided framework for self-supervised 6D pose estimation of UUVs as illustrated in \cref{fig:teaser}. 
We aim to refine an object pose estimator initially trained on synthetic data and adapt it to real-world underwater data. 
FAFA focuses on RGB-based 6D pose estimation, without additional supervision sources. 
In the pre-training stage, motivated by frequency domain image analysis \cite{oppenheim1979phase, piotrowski1982demonstration, hansen2007structural,he2023camouflaged}, we introduce a Fast Fourier Transform (FFT) based data augmentation method that involves amplitude mix and amplitude dropout. 
Specifically, by blending amplitude information from both synthetic and real images, we unveil style priors from the unlabeled target domain. 
Meanwhile, randomly dropping amplitudes aids the network in learning domain-invariant features. 
In the subsequent self-supervised refinement stage, we introduce multi-level flow-aided consistencies that enforce alignments at both the image and feature levels, thereby enhancing the network's performance in a more effective manner.
Focusing solely on image-level alignment may overlook latent representations that are crucial for semantic and geometric consistencies \cite{peng2021full}. 
Thereby, establishing feature-level alignment within a high-dimensional latent space, beyond mere image-level alignment, enables a more effective extraction and integration of high-level semantic information, making our self-supervision more effective.
%
%
%
%
%
Our contributions are summarized as follows.
\begin{itemize}
\item We present a two-stage self-supervised framework for underwater RGB-based 6D pose estimation, which effectively leverages unlabeled underwater images for end-to-end domain adaptation.

\item We introduce a Frequency-aware augmentation strategy to improve the overall pose accuracy as well as the network's adaptability to diverse underwater domains.

\item We propose to establish multi-level flow-aided consistencies encompassing both image-level and feature-level alignments, enhancing the effectiveness of self-supervision.

\end{itemize}
We assess our approach, FAFA, on underwater 6D object pose benchmarks, namely ROV6D \cite{tang2023rov6d} and DeepURL \cite{joshi2020deepurl}, and showcase its remarkable superiority over the current state-of-the-art, without reliance on any additional real-world (self-)supervision signals.

%% file: figs/0_teaser.tex
\begin{figure}[t]
\centering
\includegraphics[width=0.95\textwidth]{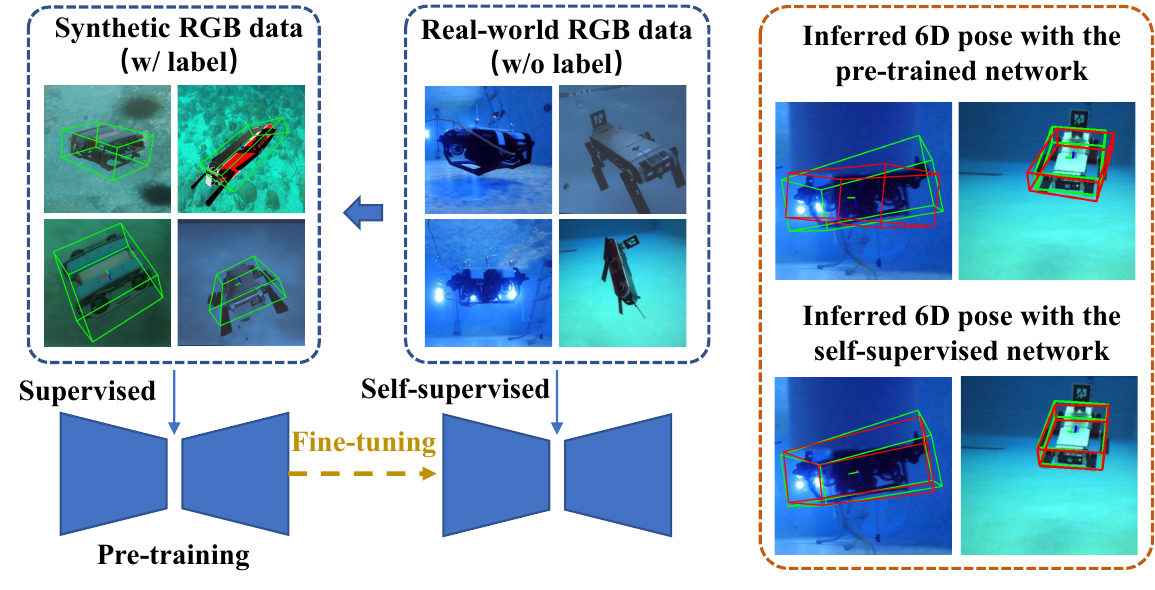}
\caption{Abstract illustration of our proposed approach. We initially train the network using annotated synthetic RGB data. Subsequently, real-world unlabeled data are employed for self-supervised learning to further refine the network. After that, The network's performance exhibits a significant improvement. The green and red bounding boxes denote the ground truth and prediction, respectively.}

\label{fig:teaser}
\end{figure}

%% file: sections/2_related.tex
\section{Related Work}
\label{sec:related}

\subsection{Underwater Object Pose Estimation}
Underwater object pose estimation, due to the inherent difficulty in acquiring depth information, closely relates to monocular 6D object pose estimation in generic scenarios which primarily involves two main categories of methods. 
The first category involves indirect methods \cite{li2019cdpn, rad2017bb8, tekin2018real, hu2019segmentation, peng2019pvnet}, which identify target keypoints and then employ a P$n$P solver to calculate the pose. 
The second category consists of direct methods \cite{xiang2017posecnn, bukschat2020efficientpose,kehl2017ssd}, which directly regress the object pose. 
However, most of these methods rely on annotated labels for training, which is challenging and impractical for underwater scenes. 
A commonly adopted approach is to use synthetic data for training, such as utilizing UE4 and CycleGAN \cite{zhu2017unpaired} to generate Aqua2 robot data \cite{joshi2020deepurl} or ROV6D for BlueROV robot data \cite{tang2023rov6d}. 
Nevertheless, a noticeable domain gap often exists between synthetic and real data. 
Some methods employ refinement strategies to fine-tune networks trained on synthetic data \cite{li2020deepim, labbe2020cosypose}, but they still require a small amount of annotated real data for further refinement. 
Achieving improved network performance without any annotated real data remains challenging.

\subsection{Self-Supervision and Domain Adaptation for Object Pose Estimation}
The issue of poor generalization performance in networks trained on synthetic data has prompted the exploration of self-supervised frameworks to address the challenges associated with obtaining annotations for real data. 
Various methods, such as \cite{wang2020self6d,deng2020self,chen2022sim,wang2021occlusion}, have been proposed to alleviate the difficulty of acquiring labeled real data. 
While several of these methods \cite{wang2020self6d,deng2020self,chen2022sim} require additional depth information during the self-supervised process, the application of depth sensors is limited in underwater scenarios.
Self6D++ \cite{wang2021occlusion} introduces crucial modifications, such as the simultaneous prediction of visible and amodal masks and the utilization of noise-aided student training, thereby mitigating the reliance on depth information. 
However, it relies on effective refinement methods for obtaining pseudo-labels, and existing refinement techniques exhibit suboptimal performance in underwater scenes.
PFC \cite{hai2023pseudo} addresses the challenge by incorporating an optical flow network to establish self-supervised pixel-wise geometric consistency. 
Nevertheless, the underlying matching process fails to take target shape constraints into account, potentially resulting in noisy pseudo flow labels and poor pose estimates. 
Furthermore, PFC is not amenable to end-to-end training, and its optimization is susceptible to the impact of P$n$P-RANSAC.

Existing self-supervised methods typically employ synthetic data for pre-training, followed by further self-supervised fine-tuning on real data. 
However, there is limited consideration for the domain gap between synthetic and real data in these approaches. 
Moreover, the naive fine-tuning methods may cause the network to converge to unfavorable local minima.


To address this issue, 
prevalent strategies involve adversarial training to generate data with the stylistic attributes of the target domain \cite{joshi2020deepurl, zhang2022self}, or facilitating feature alignment between the source and target domains \cite{zheng2023gpdan}. 
These methods impose high demands on network design and parameter tuning, making the training process challenging to converge effectively.
Domain adaptation methods based on the Fourier transform perspective have garnered attention \cite{yang2020fda, Yang_2020_CVPR, Chang_2019_CVPR, yi2021complete}, primarily because the phase and amplitude components of the Fourier spectrum encapsulate high-level semantics and low-level information of the original image, respectively. 
Exploiting this property, domain adaptation methods utilizing Fourier transform have achieved significant success in target segmentation and recognition \cite{yang2020fda, Yang_2020_CVPR, zhang2018fully}. 
However, research in the context of target pose estimation remains relatively scarce.

%% file: sections/3_method.tex
\input{figs/1_method}

\section{Approach}
\label{sec:method}

Given the 3D CAD model of an object, a set of synthetic RGB images with 6D pose annotations, and an unlabeled set of real RGB images captured underwater, our goal is to obtain accurate object poses from real underwater RGB images.
To this end, we propose FAFA, a Frequency-Aware Flow-Aided self-supervised framework leveraging a two-stage training pipeline for underwater 6D pose estimation.
This approach initially estimates a coarse object pose and further refines it through a self-supervised network. 
We introduce a frequency-based data augmentation method during the pre-training process, diverging from the previous practice of relying solely on synthetic data for training. 
Within our self-supervised framework, we employ a shape-constraint optical flow and propose multiple alignment strategies to facilitate pose refinement. The framework is implemented based on a differentiable learning pipeline.
The overview of our method is depicted in \cref{fig:method}.
\red{In the following, we first provide an overview of FAFA in \cref{sec:fafa_ss_framework}, then introduce the FFT-based data augmentation strategy in \cref{sec:FFT}. }
Finally, \cref{sec:flow_ss_pose_est} details the flow-aided multi-level self-supervision.

\subsection{Framework of FAFA}
\label{sec:fafa_ss_framework}
\red{
We employ a teacher/student architecture for our self-supervised framework, comprising two networks with identical structures and shared initial weights. The
network consists of three main components: 1) a feature encoder;
2) a flow regressor;
3) a pose regressor that estimates a relative pose $\textbf{P}_{\Delta}$ from the intermediate flow $f$ to update the previous pose $\textbf{P}_{j-1}$.
Our network then generates the shape-constraint flow based on the
current pose to iteratively update the flow field and pose, and finally output the refined pose.
}

\red{
During the pre-training stage, $\mathbf{P}_0$ is obtained using existing 6D pose estimation methods such as DeepURL \cite{joshi2020deepurl} and GDRN \cite{wang2021gdr}.
During self-supervised training, 
the real image and a set of synthetic images rendered around $\textbf{P}_0$ are input to the teacher network.
As in \cite{hai2023pseudo}, when the pixel-wise flows between the synthetic image $I^{s}$ and the real image $I^{r}$ calculated by the teacher model satisfy geometric consistency, we select them as pseudo-labels $f^{tea}$ to supervise the student network with the noisy input.
The further pose predicted by the teacher network is directly used as a pseudo pose label.
The self-supervision primarily involves three parts.
First, we conduct feature-level alignment by utilizing the flow of the student network $f^{stu}$, and the features derived from real images through its inherent feature encoder to supervise the synthesized features.
Second, we employ transformations on $I^{r}$ and the synthetic mask, anchored on $f^{tea}$, thereby supervising the transformed outputs using $f^{stu}$ for image-level regulation.
Moreover, we employ point-matching loss for aligning 6D poses.}

\input{figs/2_FFT}

\subsection{Fourier-based Perspective Augmentation}
\label{sec:FFT}
To bridge the gap between synthetic and real domains, our objective during the synthetic pre-training stage is to encourage the network to not only acquire domain-invariant features but also reveal style priors pertinent to the target domain. 
Therefore, we draw inspiration from the frequency perspective to foster these desirable properties.
The Fourier transform $\mathcal F$ can decompose an image to a phase component $\mathcal F^P$ and an amplitude component $\mathcal F^A$. 
Intuitively speaking, the phase part encompasses domain-invariant information like shapes or contours, while the amplitude component contains the domain-specific information like contents or styles \cite{oppenheim1979phase, piotrowski1982demonstration, hansen2007structural}. 
By manipulating the phases and amplitudes within the Fourier domain, we devise a data augmentation strategy aimed at facilitating the learning of both domain-invariant and target domain-specific features.

Formally, when provided with an underwater image \textit{x}, we employ the FFT algorithm \cite{nussbaumer1982fast} to extract phase and amplitude information as follows:

\begin{equation}
    \begin{aligned}
    &\mathcal{F}^A(x)(u, v)=\sqrt{R^2(\textit{x})(u, v)+\textit{I}^2(\textit{x})(u, v)} \\
    &\mathcal{F}^P(x)(u, v)=\arctan\frac{\textit{I}(\textit{x})(u, v)}{\textit{R}(\textit{x})(u, v)}
    \end{aligned},
\end{equation}
where \textit{R}(\textit{x}) and \textit{I}(\textit{x}) are the real and imaginary parts of $\mathcal F$(\textit{x}), respectively. Accordingly, $\mathcal F^{-1}$ represents the inverse Fourier transformation.

As shown in Fig.~\ref{fig:fft}, we initially extract both the amplitude and phase components, and then reconstruct the images based on each component. 
Notably, the phase-only reconstructed image exhibits clear outlines, allowing for a quick and straightforward identification of the object. Nevertheless, recognizing the object through amplitude-only reconstruction is unfeasible.
Motivated by \cite{yang2020fda}, the phase containing contours and edges information is not sensitive to domain shifts and the amplitude reveals the image style. 
Based on this property, we randomly discard amplitude spectra to facilitate the network to learn the domain-invariant phase information. On the other hand, we perform linear interpolation on the amplitude spectra of two images from the source and target domains to generate a new amplitude spectrum. 
The goal is to introduce style priors of the target domain for training with synthetic data.
Concretely, given two images $x^s$ and $x^r$ randomly sampled from synthetic and real-world datasets, the transformed amplitude information is defined as:
\begin{equation}
    \hat{\mathcal{F}}^A(x^s, x^r) =\left\{
    \begin{aligned}
    &(1-\alpha){\mathcal{F}}^A(x^s) + \alpha {\mathcal{F}}^A(x^r), &\delta<\delta_0 \\
    &1, &\text{otherwise} \\
    \end{aligned}
    \right. 
\end{equation}
where the hyperparameter $\delta_0$ $\in$ (0,1) controls the dropout of the amplitude spectra and $\alpha$ $\sim$ \textit{U}(0, $\beta$). 
The parameter $\beta$ reflects the augmentation strength.

Finally, the transformed amplitude information and the original phase information are reconstructed to generate the augmented image \red{$x^{s \rightarrow r}$} for training:
\begin{equation}
\red{x^{s \rightarrow r}}= \mathcal F^{-1} [\hat{\mathcal F}^A(x^s, x^r)(u, v) \ast e^{-j \ast {\mathcal F}^P(x^s)(u, v)}].
\end{equation}




\subsection{Flow-aided Self-supervised 6D Pose Estimation}
\label{sec:flow_ss_pose_est}
In this section, we propose a flow-aided self-supervised framework. 
Inspired by \cite{hai2023shape}, we build our object pose estimator on top of a flow-aided direct pose regression network, wherein iterative optimization is performed for both the pose and optical flow. 
Specifically, shape constraints based on the object pose are imposed during flow estimation, and then we can learn a more accurate target pose by the refined flow in an end-to-end manner. 
During self-supervised learning, unlike \cite{hai2023pseudo}, which solely focuses on image-level constraints within the teacher-student network, we introduce diverse flow-aided self-supervised consistencies to establish comprehensive multi-level alignments, encompassing both the image level and feature level.

\nbf{Flow-aided Object Pose Estimator.}
Given an input real image $I^r$ and an initial pose $\mathbf{P}_0$ for a target object, we first use its 3D model $\mathcal{M}$ to render $N$ synthetic images $\{I^s_i\}_{i=1}^N$ around $\mathbf{P}_0$. For the flow-aided object pose estimator, we establish the dense 2D-2D correspondences between real and synthetic images $\{(\textit{$I^r$},\textit{$I^s_i$})\}_{i=1}^N$ to fine-tune the pose $\textbf{P}_0$, which has been shown promising for self-supervised frameworks \cite{hai2023pseudo}. 
Given two pixel locations $x^t$ and $x^s$ of matching points in \textit{$I^r$} and \textit{$I^s_i$}, in contrast to \cite{hai2023pseudo}, we propose to let the network predict the shape-constraint flow:
\begin{equation}
    x^t_i = x^s_i + \sum_{j=1}^{S}f^{s \rightarrow t}_i(\mathbf{P}_{j-1},\mathcal{M}, K),
\end{equation}
where $f^{s \rightarrow t}$ represents the dense 2D-2D flow vector, $S$ is the number of iterations,
\red{$\textbf{P}_{j-1}$ is the intermediate predicted pose extracted from the previous flow field at iteration $j$ and $K$ is the camera intrinsic matrix. }
Let $M_{i}$ be the object visible mask for the observed RGB image.
To generate the pose-guided flow $f^{s \rightarrow t}$, we calculate the 2D-3D correspondences between the input image and the 3D object points in each iteration. 
Specifically, the correspondence \{$p_i\rightarrow\textbf{u}_i$\} of point $p_i$ of 3D mesh model to the 2D location $\textbf{u}_i=({u}_i,{v}_i)$ can be obtained by:

\begin{equation}
    \forall \textbf{u}_i \in M_{i}, \lambda_i
    \left[\begin{array}{c}
    \textbf{u}_i\\
    1
    \end{array} 
    \right]
    = K(\textbf{R}_jp_i+\textbf{t}_j),
\end{equation}
where $\lambda_i$ is a scale factor. 
$\textbf{R}_j$ and $\textbf{t}_j$ are the rotation and translation of pose $\textbf{P}_j$. 
Then we estimate the flow field $\{f^{s \rightarrow t}_i=\textbf{u}^{s}_{ij}-\textbf{u}^{r}_{i0}\}$, and $\textbf{u}^r_{i0}$ can be obtained with the initial pose $\textbf{P}_0$. 
Similar to \cite{li2020deepim}, a decoupled relative pose [$\textbf{R}_{\Delta}$|$\textbf{t}_{\Delta}$] is predicted and iteratively refined based on an initial pose. 
By decoupling the estimation of rotation from translation, we effectively mitigate significant errors associated with the intricate SE(3) space. 
Instead of utilizing quaternions or Euler angles, we employ a continuous 6-dimensional rotation representation \cite{wang2021gdr, zhou2019continuity} to facilitate the direct regression of pose within a neural network.






\nbf{Image-level Alignment for Self-supervision.}
For image-level alignment, we formulate a learning objective based on the predicted flow field ${f}^{stu}$ and our pseudo flow field ${f}^{tea}$ that consists of forward-backward flow consistency and self-supervision in optical flow. 

Firstly, our goal is 
to align real images that have been backward-warped using the predicted flow field ${f}^{stu}$ with those that have been transformed by the pseudo flow field ${f}^{tea}$.
We utilize the photometric consistency $\mathcal{L}_{photo}$ to measure the appearance similarity, which is defined by the difference between the warped real images $\widetilde{\omega}$($I^r,{f}^{stu}$) and $\widetilde{\omega}$($I^r,{f}^{tea}$):
\begin{equation}
    \mathcal{L}_{photo} = \sum_{i=1}^{N}M_i \rho(\widetilde{\omega}(I^r,{f}_i^{stu}), \widetilde{\omega}(I^r,{f}_i^{tea})).
\end{equation}
Here $\omega$ and $\widetilde{\omega}$ denote the forward and backward warping transformations based on the flow field, and $\rho$ is the Census loss \cite{meister2018unflow,jonschkowski2020matters}. 
For simplicity, we only calculate the optical flow of 3D points on the visible surface. 
Moreover, we evaluate the consistency of the forward flow on the warped rendered masks according to:
\begin{equation}
    \mathcal L_{warp-mask} = \sum_{i=1}^{N} \Vert \omega(M_i^s,{f}_i^{stu}), \omega(M_i^s,{f}_i^{tea}) \Vert_{1},
\end{equation}
where $M^s$ is the mask of the rendered images. Furthermore, we employ the pseudo flow field ${f}^{tea}$ to supervise the flow field ${f}^{stu}$ predicted by the student network, utilizing $\mathcal{L}_{flow}$ as described in \cite{hai2023pseudo}. 
The image-level alignment is then composed as the weighted sum over all three terms:

\begin{equation}
    \mathcal{L}_{img-level} = \gamma_1 \mathcal{L}_{flow}+\gamma_2 \mathcal{L}_{photo}+ \mathcal{L}_{warp-mask},
\end{equation}
where $\gamma_1$ and $\gamma_2$ are the associated loss weights.

\nbf{Feature-level Alignment for Self-supervision.}
Compared to optical flow estimation for indoor RGB images, there is a significant disparity between real images and synthetic images in underwater environments.
These differences primarily manifest in appearance and derive from environmental variations, such as changes in color, lighting, and clarity. 
Inspired by \cite{lee2024semi}, we introduce a feature-level error that exhibits robustness to environmental changes, instead of relying solely on image-level errors.
The features $F^r$ and $F^s$ are extracted by the feature encoder of the pose network for a real underwater image and a synthetic image, respectively. 
The proposed feature-level metric $\mathcal{L}_{feat-level}$ is defined as the weighted average of the feature dissimilarity between the $F^r$ warped by $f^{stu}$ and $F^s$.

\nbf{Overall Self-supervision.}
In works for direct pose regression, the pose loss is also crucial for optimization. 
Thus we utilize a commonly used point matching loss $\mathcal{L}_{pose}$ \cite{xiang2017posecnn, wang2021gdr, hai2023shape} to align the 6D pose. 

Eventually, our self-supervised loss can be summarized as:

\begin{equation}
    \mathcal{L}_{self} = \gamma_3 \mathcal{L}_{pose}+\gamma_4 \mathcal{L}_{img-level}+ \mathcal{L}_{feat-level},
\end{equation}
where $\gamma_3$ and $\gamma_4$ are the corresponding loss weights.




%% file: figs/1_method.tex
\begin{figure}[t]
\centering
\includegraphics[width=0.95\textwidth]{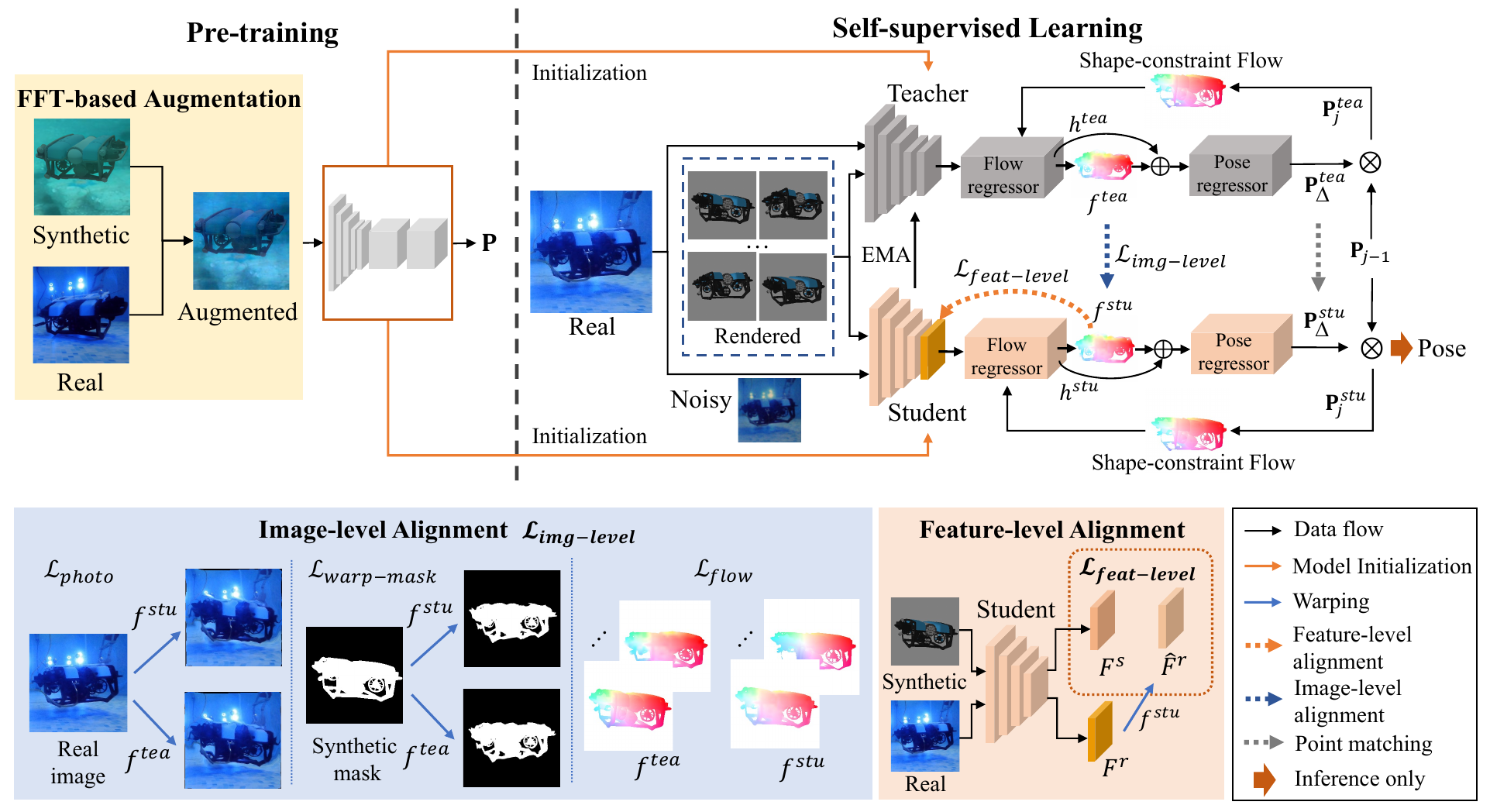}
\caption{Two-stage self-supervised framework for underwater object pose estimation. Top: 
We introduce an FFT-based data augmentation strategy, leveraging random real-world images to generate the augmented ones. 
We initialize the teacher-student framework based on the pre-trained network. The real image, along with a set of synthetic images generated based on $\textbf{P}_0$, are input to the self-supervised network. The teacher/student network consists of three components: (1) A feature encoder. (2) A flow regressor \cite{teed2020raft} that outputs a hidden feature $h$ and further estimates a flow field $f$. (3) A pose regressor which predicts a relative pose $\textbf{P}_{\Delta}$ to generate a shape-constraint flow field. 
The shape-constraint flow is then feedback to the flow regressor for iterative network optimization. Finally, the refined pose is output.
During self-supervision, the results ($f^{stu}$, $\textbf{P}^{stu}$) estimated from the noisy inputs are supervised by pseudo-labels ($f^{tea}$, $\textbf{P}^{tea}$) obtained from clean images.  Bottom: We optimize flow and pose estimation by simultaneously applying image-level and feature-level alignment constraints.}
\label{fig:method}
\end{figure}

%% file: figs/2_FFT.tex
\begin{figure}[t]
\centering
\includegraphics[width=0.75\textwidth]{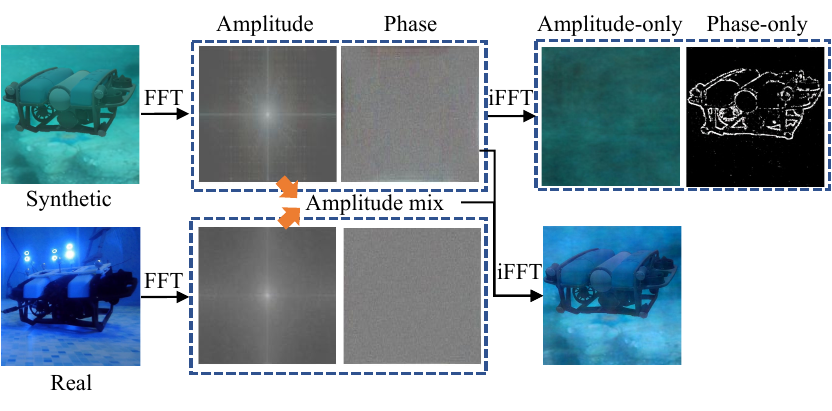}
\caption{FFT-based augmentation strategy. 
Top: The FFT algorithm extracts amplitude and phase components from a synthetic image. 
Images are then reconstructed by exclusively utilizing either the amplitude or the phase component through the inverse Fourier transform (iFFT). Bottom: Real image amplitude information is introduced, and a new blended amplitude component is formed by mixing it with the synthetic image amplitude. 
Subsequently, an augmented image is reconstructed by combining this new amplitude component with the synthetic image's phase component.}
\label{fig:fft}
\end{figure}

%% file: sections/4_exp.tex
\section{Experiment}


\subsection{Experimental Setup}
\label{sec:exp_setup}

\nbf{Datasets.}
We evaluate our method on two datasets: ROV6D \cite{tang2023rov6d} and DeepURL \cite{joshi2020deepurl}. 
\emph{ROV6D} \cite{tang2023rov6d} is a large-scale underwater dataset with 6D pose annotations, focusing on estimating the 6D pose of a BlueROV. 
The dataset contains a synthetic subset for training and two real-world subsets collected in both a pool and the open lake at Maoming for testing. 
The Pool subset is further divided into the Pool-Basic group and the Pool-Occluded group, where the object in the Pool-Occluded group is partially occluded by different obstacles. 
The Maoming subset, however, contains only a small amount of data with low visibility. 
Limited by the number of images in the Maoming subset, we exclusively employ the Pool subset for evaluation. 
For self-supervised training, we utilize 1000 images of the Pool-Basic group data. 
Subsequently, the Pool-Occluded group is employed for testing and evaluating the model's performance. 
\emph{DeepURL} \cite{joshi2020deepurl} focuses on estimating the 6D pose of the Aqua2 robot, collecting a Pool dataset in both indoor and outdoor real-world underwater environments. 
The training dataset comprises a large number of synthetic images generated using UE4 and CycleGAN. 
To fully evaluate our data augmentation approach, we exploit BlenderProc \cite{denninger2020blenderproc}, a Blender-based rendering pipeline for synthesizing data, to generate 30,000 synthetic images with GT 6D poses for Aqua2. 
Note that the target real-world backgrounds in the testing subsets are not utilized in the training data generation process.
Similar to ROV6D, we randomly selected 1000 real images for self-supervised training, with the rest real images employed for evaluation. \red{In addition, to show that FAFA is also effective on general scenarios beyond underwater applications, we present the results on YCB-V \cite{xiang2017posecnn} in the Appendix.}

\nbf{Metrics.}
We evaluate our method using two standard metrics: 
ADD(-S)~\cite{hinterstoisser2013model,hodavn2016evaluation} and \textit{n}° \textit{n} cm~\cite{6619221}. 
ADD(-S) is utilized to compute the 3D Euclidean averaged distance of all model points between the estimated pose and the ground truth pose. 
For ADD, the estimated pose is considered to be correct if the average distance is below 10\% of the model diameter d. 
The ADD-S metric is employed for symmetric objects to measure the average distance to the closest model point.
Following \cite{tang2023rov6d}, we regard BlueROV as a symmetric object and utilize ADD-S for evaluation on ROV6D.
For metric \textit{n}° \textit{n} cm, it considers an estimated pose to be correct if its rotation error is within \textit{n}° and the translation error is below \textit{n} cm. 
To distinguish their impact, we evaluate them separately for \textit{n}° and \textit{n} cm.


\nbf{Implementation Details.}
We implement our method with PyTorch \cite{paszke2019pytorch} and conduct all experiments on an NVIDIA RTX-3090 GPU. 
Similar to \cite{hai2023pseudo}, our network is trained using the AdamW optimizer \cite{loshchilov2017decoupled} with a batch size of 16 and the OneCycle learning rate schedule \cite{smith2019super} with a maximum learning rate of $4\times10^{-4}$. 
The proposed FFT-based augmentation strategy is only utilized during the pre-training process.
Before self-supervised training, we crop the input based on the initial pose and resize the image patch to 256$\times$256. 
We use $N =$ 4, $\delta_0 =$ 0.5, $\beta =$ 1 in our work. 
The number of iterations $S$ is set to 8 for the teacher network and 4 for the student network, respectively. 
The weights of the teacher network are updated from the student network by a simple exponential moving averaging (EMA) with an exponential factor of 0.999 \cite{tarvainen2017mean}. 
The self-supervision hyper-parameters are configured as follows: 
$\gamma_1 =$ 0.1, $\gamma_2 =$ 0.1, $\gamma_3 =$ 10 and $\gamma_4 =$ 10.
Given the distribution disparities between synthetic and real data, we additionally freeze batch normalization within the network to enhance the stability of the self-supervision.

\subsection{Comparison with State-of-the-art Methods}
\label{sec:exp_sota}

We compare our approach with state-of-the-art methods on the ROV6D and DeepURL datasets. 
We re-implement PFC \cite{hai2023pseudo} and evaluate it on the two datasets. 
Our method and PFC employ the same 1000 real images for training during self-supervision.
For better comprehension, in addition to presenting the results of our self-supervised model, we evaluate our approach using a pre-trained model and a model trained under supervision with labeled real-world underwater images.
These settings respectively serve as lower and upper bounds for our method, and we denote them as OURS (LB) and OURS (UB) below.

\nbf{Results on the ROV6D Dataset.}
\input{tables/result_rov6d}
\cref{tab:result_rov6d} presents the results of our method compared with state-of-the-art methods \cite{li2019cdpn, wang2021gdr, li2020deepim, hai2023shape, hai2023pseudo} on ROV6D. 
We categorize the results into groups based on whether the methods support self-supervised training on unlabeled real data. 
Note that even though the real data utilized has accurate annotations in our experiments, no annotations are used during the training process except OURS (UB). 
For instances where self-supervision is not conducted, our model demonstrates state-of-the-art performance compared to other methods. 
Referring to \cref{tab:result_rov6d}, OURS (LB) achieves a significant improvement with the ADD-S 0.1d metric of 93.54\%, the 5° metric of 78.93\% and the 5 cm metric of 80.10\%, outperforming flow-based method SCFlow \cite{hai2023shape}, as well as the dense correspondence-based methods CDPN \cite{li2019cdpn} and GDRN \cite{wang2021gdr}. 
DeepIM \cite{li2020deepim} is a refinement method that utilizes optical flow estimation as an auxiliary task. 
However, its performance is significantly limited, especially in underwater challenging scenarios.

On the other hand, in cases involving training with unlabeled real data, OURS reveals the 5° metric and 5 cm metric of 87.87\% and 86.71\%, respectively, 
which is better than the previous state-of-the-art self-supervised method PFC \cite{hai2023pseudo}  reporting 81.34\% and 72.10\%.
Surprisingly, the results after self-supervised training are comparable to OURS (UB), with an ADD-S 0.1d score of 96.96\%, even surpassing OURS (UB) of 96.21\%. These results demonstrate the robustness of our method in handling occlusions. Fig.~\ref{fig:exp} (a) illustrates some qualitative results on the Pool-Occluded group of the ROV6D dataset.

\input{figs/3_exp}

\nbf{Results on the DeepURL Dataset.}
\input{tables/result_deepurl}
We compare our approach to the methods \cite{joshi2020deepurl, wang2021gdr, li2020deepim, hai2023shape, hai2023pseudo} on DeepURL in \cref{tab:result_deepurl}. 
During the pre-training stage, our method introduces frequency-aware augmentation to synthetic data. 
To avoid potential confusion with the effects of CycleGAN style transfer, our approach, along with the self-supervised network PFC, is trained using data generated by PBR. 
In contrast, methods without a self-supervised stage use synthetic data that undergoes style transfer via CycleGAN. 
OURS (LB) is only slightly inferior to DeepIM~\cite{li2020deepim}. 
Furthermore, our approach and PFC surpass OURS (UB) under the ADD 0.1d metric, demonstrating a clear advantage of self-supervised learning. 
Our method outperforms PFC in overall performance, achieving a 5° metric of 68.70\% and a 5 cm metric of 50.92\%, compared to PFC reporting 61.60\% and 44.10\%. 
Fig.~\ref{fig:exp} (b) shows some qualitative results.

\subsection{Ablation Study}
\label{sec:exp_ablate}

\cref{tab:ablation_deepurl} and \cref{tab:ablation_diff_comp} present several ablations on DeepURL \cite{joshi2020deepurl}.
First, we train our model using PBR synthetic data. 
Then, we leverage 1000 real-world images for self-supervised training. 
\red{Note that for evaluation, no data from the training stages is utilized, including pre-training and self-supervision.}

\input{tables/exp_ablation}

\nbf{Evaluation of FFT-based Data Augmentation Strategy.}
As mentioned in \cref{sec:FFT}, the amplitude mix operation introduces the style priors of the target domain for synthetic data, while amplitude dropout promotes the learning of domain-invariant features.
\cref{tab:ablation_deepurl} (B0-B3) presents the performance regarding four different data augmentation settings for pre-training, specifically involving the removal of amplitude mix or amplitude dropout operations. 
Yet, we surpass our baseline without utilizing FFT-based data augmentation operations (B0 \vs B3). 
After self-supervised training, the results employing any augmentation strategy during the pre-training stage outperform PFC (A0 \vs C0-C2). \red{In addition, employing FFT augmentation on real images may hinder capturing the true distribution of the target domain (C0 \vs C3).}
Our method achieves a significant improvement of 15.46\% under the 5cm metric, increasing from 44.10\% to 50.92\% (A0 \vs C0).
These results demonstrate that our data augmentation strategy enhances the performance of networks trained with synthetic data when applied to real-world scenarios.




\nbf{Evaluation of Different Components.}
We ablate different components, including the image-level alignment terms $\mathcal{L}_{photo}$, \red{$\mathcal{L}_{warp-mask}$} and the feature-level alignment term $\mathcal{L}_{feat-level}$ in \cref{tab:ablation_diff_comp}. 
The first row shows the results when considering only $\mathcal{L}_{flow}$ and $\mathcal{L}_{pose}$, as used in SCFlow \cite{hai2023shape}. 
After employing our image-level alignment term $\mathcal{L}_{photo}$, the result outperforms the baseline by a margin of 23.28\% in terms of 5° metric. 
Building upon this improvement, the network equipped with \red{$\mathcal{L}_{warp-mask}$} produces more accurate results, with the 5cm metric increasing from 43.90\% to 49.36\%. 
Our method attains its best performance when both image-level and feature-level losses are simultaneously employed.
These results demonstrate the effectiveness of our proposed components.

\input{tables/ablation_diff_comp}



%% file: tables/result_rov6d.tex
\begin{table}[t]
  \centering
   \caption{Results on the ROV6D dataset. All methods utilize synthetic data generated by PBR for training. PFC and Our method further leverage 1000 real images for self-supervised training. OURS (LB) only employs real data for image augmentation. The best label-free method is highlighted in \textbf{bold}, and the overall best method is \underline{underlined}.
  }
  \label{tab:result_rov6d}
  \tablestyle{8pt}{1}
  \begin{tabular}{@{}l|c|cc|c|c|c@{}} 
    \multirow{2}*{Method} & \multirow{2}*{Real Data} &\multicolumn{2}{c|}{ADD-S} &\multirow{2}*{5°}&\multirow{2}*{5\,cm}&\multirow{2}*{MEAN}\\
    & &0.05d&0.1d&&\\
    \shline
    CDPN \cite{li2019cdpn} & - &81.56 &88.43&53.21&56.44&69.91\\
    GDRN \cite{wang2021gdr}& -&89.64&91.74&57.12&73.08&77.90\\
    DeepIM \cite{li2020deepim}& - &60.05&76.76&12.96&58.32&52.02\\
    SCFlow \cite{hai2023shape}& - &82.09 &90.99&73.34&71.46&79.47\\
    OURS (LB) & Unlabeled &87.68 &93.54&78.93&80.10&85.06\\
    \hline
    PFC \cite{hai2023pseudo} & Unlabeled &86.56 &93.50&81.34&72.10&83.38\\
    OURS & Unlabeled & \textbf{94.03} &\textbf{\underline{96.96}}&\textbf{87.87}&\textbf{86.71}&\textbf{91.39}\\
    \hline
    OURS (UB) & Labeled & \underline{95.08} &96.21&\underline{88.81}&\underline{89.60} & \underline{92.43} \\

  \end{tabular}


\end{table}

%% file: figs/3_exp.tex
\begin{figure}[t]
\centering
\includegraphics[width=0.95\textwidth]{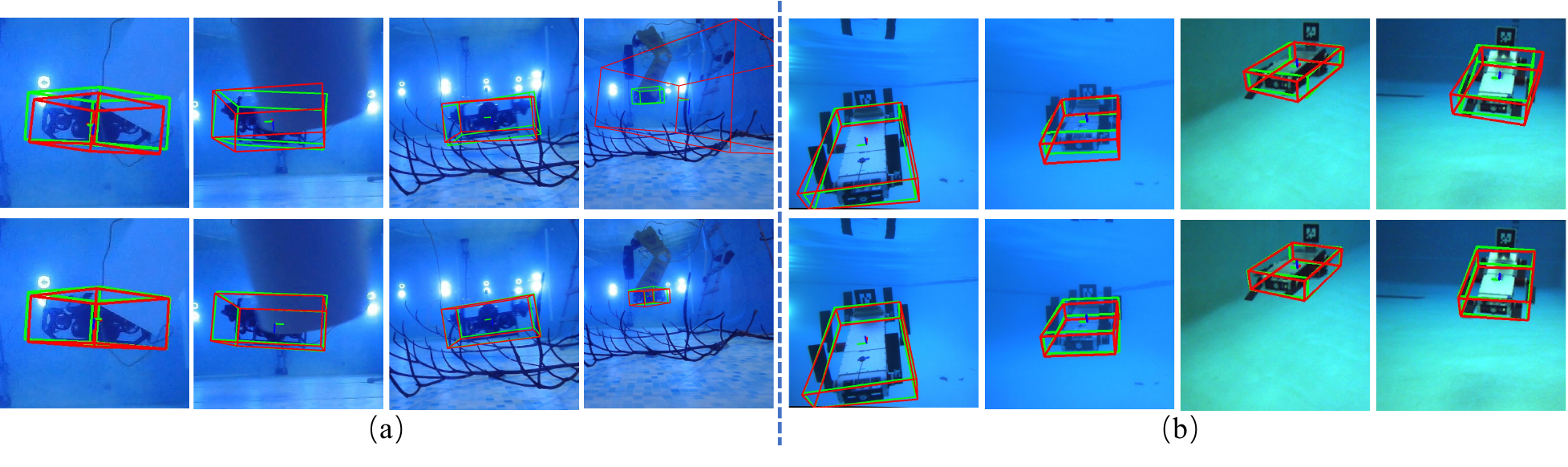}
\caption{Qualitative results on (a) ROV6D and (b) DeepURL. The results are obtained before (top) and after (bottom) employing our self-supervision, respectively. The green and red wireframes represent the ground-truth pose and the results.}
\label{fig:exp}
\end{figure}

%% file: tables/result_deepurl.tex
\begin{table}[t]
  \centering
  \caption{Results on DeepURL dataset. The synthetic data has two categories: (1) one generated by UE4 and CycleGAN with the target domain style, and (2) the other created by PBR, independent of the target domain.
  After pre-training, PFC and our method further leverage 1000 real images for self-supervised training. OURS (LB) only utilizes real data for image augmentation. The best label-free method is highlighted in \textbf{bold}, and the overall best method is \underline{underlined}.
  }
  \label{tab:result_deepurl}
  \tablestyle{3pt}{1}
  \begin{tabular}{@{}l|c|c|cc|c|c|c@{}} 
    \multirow{2}*{Method} &\multirow{2}*{Synthetic Data}&\multirow{2}*{Real Data} &\multicolumn{2}{c|}{ADD} &\multirow{2}*{5°}&\multirow{2}*{5cm}&\multirow{2}*{MEAN}\\
    &&&0.05d&0.1d&&\\
    \shline
    DeepURL \cite{joshi2020deepurl} &UE4+CycleGAN &-&8.69&	55.85&47.17&35.66&36.84\\
    GDRN \cite{wang2021gdr}&UE4+CycleGAN &-&13.42&55.01&30.73&42.56&37.47\\
    DeepIM\cite{li2020deepim}&UE4+CycleGAN & - &15.81&58.88&35.03&43.06&44.67\\
    SCFlow \cite{hai2023shape}&UE4+CycleGAN &-&15.66&59.41&44.65&37.75&39.37\\
    OURS (LB) &PBR & Unlabeled &18.62&73.36&48.10&38.50&44.64\\
    \hline
    PFC \cite{hai2023pseudo} & PBR & Unlabeled &70.30&\textbf{\underline{98.87}}&\red{61.60}&44.10&68.72\\
    OURS &PBR & Unlabeled &\textbf{76.56}&97.64&\textbf{68.70}&\textbf{50.92}&\textbf{73.46}\\
    \hline
    OURS (UB) &PBR & Labeled &\underline{85.60}&97.42&\underline{95.24}&\underline{92.52}&\underline{92.68}\\

  \end{tabular}
  
\end{table}

%% file: tables/exp_ablation.tex
\begin{table}[tb]
  \centering
  \caption{Ablation study of FFT-based augmentation on DeepURL. We compare our method with networks that exclude amplitude dropout, amplitude mix operations, and any FFT-based augmentation strategies during the pre-training stage. Furthermore, we conduct self-supervised training on these pre-trained networks individually. The best label-free method is highlighted in \textbf{bold}, and the overall best method is \underline{underlined}.
  }
  \label{tab:ablation_deepurl}
  \scalebox{0.96}{
  \tablestyle{0.5pt}{1}
  \begin{tabular}{@{}l|l|cc|c|c|c@{}} 
    \multirow{2}*{Row} &\multirow{2}*{Method} &\multicolumn{2}{c|}{ADD} &\multirow{2}*{5°}&\multirow{2}*{5cm}&\multirow{2}*{MEAN}\\
    &&0.05d&0.1d&&\\
    \shline
    A0  & PFC &70.30 &\textbf{\underline{98.87}}&61.60&44.10&68.72\\
    \hline
    B0 & OURS (LB) &18.62 &73.36&48.10&38.50&44.64\\
    B1 & B0: Pre-training $\rightarrow$ w/o amplitude dropout &10.83 &55.03&52.49&37.90&39.06\\
    B2 & B0: Pre-training $\rightarrow$ w/o amplitude mix &11.44 &58.54&45.03&34.80&37.45\\
    B3 & B0: Pre-training $\rightarrow$ w/o FFT-based augmentation &8.35 &50.18&44.46&37.42&35.10\\
    \hline
    C0 & OURS  &\textbf{76.56} &97.64&68.70&\textbf{50.92}&\textbf{73.46}\\
    C1 & C0: $\rightarrow$ Pre-training with B1&74.24 &98.20&62.71&46.79&70.49\\
    C2 & C0: $\rightarrow$ Pre-training with B2& 72.30&97.48&\textbf{68.93}&47.25&71.49\\
    C3 & C0: Self-supervision w/ FFT-based augmentation& 74.06&96.76&50.12&49.88&67.71\\
    \hline
    D0 & OURS (UB)  &\underline{85.60} &97.42&\underline{95.24}&\underline{92.52}&\underline{92.68}\\
  \end{tabular}
  }
\end{table}


%% file: tables/ablation_diff_comp.tex
\begin{table}[t]
  \centering
   \caption{Ablation study of different self-supervision loss components on DeepURL. We evaluate three key components of our method, including $\mathcal{L}_{photo}$, $\mathcal{L}_{warp-mask}$, and $\mathcal{L}_{feat-level}$. The first row shows the baseline, which only computes $\mathcal{L}_{flow}$ and $\mathcal{L}_{pose}$.
  }
  \label{tab:ablation_diff_comp}
  \tablestyle{6pt}{1}
  \begin{tabular}{@{}ccc|cc|c|c|c@{}} 
    \multicolumn{3}{c|}{Method} &\multicolumn{2}{c|}{ADD}&\multirow{2}*{5°}&\multirow{2}*{5cm}&\multirow{2}*{MEAN} \\
    
    $\mathcal{L}_{photo}$&$\mathcal{L}_{warp-mask}$&$\mathcal{L}_{feat-level}$&0.05d&0.1d&&&\\
    \shline
    - & - & - &69.88&97.00&50.31&39.97&64.29\\
    \checkmark & - & - &72.98&\textbf{98.73}&62.02&43.90&69.41\\
    \checkmark & \checkmark & - &71.62&98.64&63.10&49.36&70.68\\
    - & - & \checkmark &72.10&95.73&53.86&45.38&66.77\\
    \checkmark & \checkmark & \checkmark &\textbf{76.56}&97.64&\textbf{68.70}&\textbf{50.92}&\textbf{73.46}\\
  \end{tabular}
 
\end{table}

%% file: sections/10_conclusion.tex
\section{Conclusion}
We have proposed FAFA, a framework for self-supervised underwater object pose estimation, aiming to learn information from unlabeled real-world images. 
In essence, we present an FFT-based data augmentation strategy to introduce the target domain style into synthetic data during the pre-training stage and promote the network to learn domain-invariant features. 
Furthermore, in our self-supervised network, we utilize a shape-constrained optical flow and propose various image-level and feature-level alignments for improving the network's robustness to the complex underwater environment, thus achieving more accurate pose estimation. 
Our method is evaluated on two underwater UUV datasets with different styles, demonstrating superior performance compared to previous state-of-the-art methods.
\textbf{Limitations} are discussed in the Appendix.


\section*{Acknowledgements}
We sincerely thank the anonymous reviewers and area chairs for their constructive feedback on our work. This work was supported 
in part by the STI 2030-Major Projects under Grant 2021ZD0201404, in part by the National Natural Science Foundation of China
under Grant 62027826, and in part by the Shuimu-Zhiyuan Tsinghua Scholar Program.